\documentclass{article}

\PassOptionsToPackage{numbers, compress}{natbib}
\usepackage[preprint]{neurips_2022}

\usepackage[utf8]{inputenc} % allow utf-8 input
\usepackage[T1]{fontenc}    % use 8-bit T1 fonts
\usepackage{hyperref}       % hyperlinks
\usepackage{url}            % simple URL typesetting
\usepackage{booktabs}       % professional-quality tables
\usepackage{amsfonts}       % blackboard math symbols
\usepackage{nicefrac}       % compact symbols for 1/2, etc.
\usepackage{microtype}      % microtypography
\usepackage{xcolor}         % colors

\usepackage{microtype}
\usepackage[pdftex]{graphicx}
\usepackage{subfigure}
\usepackage{booktabs} % for professional tables

\usepackage{amsmath}
\usepackage{amsthm}
\usepackage{amssymb}

\usepackage{algorithmic}
\usepackage{threeparttable}
\usepackage{multirow}

\usepackage[ruled,vlined,linesnumbered]{algorithm2e}
\SetKwComment{Comment}{$\triangleright$\ }{}

\def\eqref#1{Equation~\ref{#1}}			% reference to equation
\title{Exploiting High Performance Spiking Neural Networks with Efficient Spiking Patterns}

\author{%
	{Guobin Shen$^{1, 3}$\footnotemark[1], \ Dongcheng Zhao$^{1}$\footnotemark[1], \ Yi Zeng$^{1, 2, 3, 4}$}\footnotemark[2]\\
	$^1$ Brain-inspired Cognitive Intelligence Lab, CASIA\\ 
      $^2$ Center for Excellence in Brain Science and Intelligence Technology, CAS\\
      $^3$ School of Future Technology, University of Chinese Academy of Sciences \\
      $^4$ School of Artificial Intelligence, University of Chinese Academy of Sciences \\
	\texttt{\{shenguobin2021, zhaodongcheng2016, yi.zeng\}@ia.ac.cn}
}

\begin{document}

\maketitle

\footnotetext[1]{These authors contributed equally.}
\footnotetext[2]{Corresponding Author.}

\begin{abstract}
      Spiking Neural Networks (SNNs) use discrete spike sequences to transmit information, which significantly mimics the information transmission of the  brain. Although this binarized form of representation dramatically enhances the energy efficiency and robustness of SNNs, it also leaves a large gap between the performance of SNNs and Artificial Neural Networks based on real values. There are many different spike patterns in the brain, and the dynamic synergy of these spike patterns greatly enriches the representation capability. Inspired by spike patterns in biological neurons, this paper introduces the dynamic Burst pattern and designs the Leaky Integrate and Fire or Burst (LIFB) neuron that can make a trade-off between short-time performance and dynamic temporal performance from the perspective of network information capacity. LIFB neuron exhibits three modes, resting, Regular spike, and Burst spike. The burst density of the neuron can be adaptively adjusted, which significantly enriches the characterization capability. We also propose a decoupling method that can losslessly decouple LIFB neurons into equivalent LIF neurons, which demonstrates that LIFB neurons can be efficiently implemented on neuromorphic hardware. We conducted experiments on the static datasets CIFAR10, CIFAR100, and ImageNet, which showed that we greatly improved the performance of the SNNs while significantly reducing the network latency. We also conducted experiments on neuromorphic datasets DVS-CIFAR10 and NCALTECH101 and showed that we achieved state-of-the-art with a small network structure.
\end{abstract}
\section{Introduction}

The spiking neural networks (SNNs) use discrete spike sequences to convey information, which is more consistent with how the brain processes information. Although the binarized sequences bring high energy efficiency~\cite{shen2022backpropagation} and robustness~\cite{zhao2022spiking}, they also reduce the representation ability of the spiking neural networks. The non-differential nature of the spikes also makes it challenging to apply the backpropagation algorithm directly to the training of SNNs. Therefore, training high-performance SNNs has been a pressing problem for researchers.

In addition to the conversion-based method~\cite{bu2021optimal,li2021bsnn}, which converts the well-trained deep neural networks into SNNs, the proposal of surrogate gradient makes it possible to train a high-performance SNNs~\cite{wu2018spatio,zhang2020temporal}. Researchers have tried to close the performance gap in several ways. Some researchers have borrowed mature techniques from deep learning and applied techniques such as normalization~\cite{kim2020revisiting,zheng2021going,wu2019direct,xu2021exploiting} and attention~\cite{zhu2022tcja,yao2021temporal,yao2022attention}, etc. to the training of SNNs. This greatly improved the performance of SNNs but ignored the characteristics of SNNs. Some researchers have tried to improve and enhance the learning ability of SNNs structurally by borrowing more complex connections in the brain. BackEISNN~\cite{zhao2022backeisnn} took inspiration from the autapses in the brain and introduced the self-feedback connection to regulate the precision of the spikes. LISNN~\cite{cheng2020lisnn} modeled the lateral interactions between the neurons and greatly improved the performance and robustness. However, these methods have improved the learning ability of SNNs to some extent, but they are still far from artificial neural networks (ANNs).

Spiking neurons have rich spatio-temporal dynamics and are highly capable of information processing. ~\cite{beniaguev2021single} found that it takes a multilayer neural network to simulate the complexity of a single biological neuron. Realizing the computational power of spiking neurons, researchers tried to build more adaptive neurons.  \cite{chen2021deep} introduced the neural oscillation and spike-phase information to construct a resonate spiking neuron. \cite{fang2021incorporating,zhang2021skip} introduced the learnable time constant of the spiking neurons to boost the performance of SNNs on different challenging tasks. \cite{yin2021accurate, rathi2021diet} introduced an adaptive threshold mechanism to control the firing of spiking neurons. These works have greatly enriched the dynamics of spiking neurons, but the binarized representation creates a performance gap between them and the float-based ANNs.

Other researchers tried to design a better surrogate gradient function to reduce the information mismatch caused by inaccurate gradients in backpropagation. \cite{chen2022gradual} proposed a gradual surrogate gradient learning algorithm to ensure the precision as well as the effectiveness of the gradient during backpropagation. \cite{yin2021accurate} proposed activity-regularizing surrogate gradients, which exceeded the state-of-the-art performance for SNNs on the challenging temporal benchmarks. \cite{li2021differentiable} introduced the adaptively evolved Differentiable Spike functions to find the optimal shape and smoothness for gradient estimation based on their finite difference gradients. However, the binarized information transfer method still limits the representation ability of SNN.

As a result, some researchers try to enrich the representation ability of SNNs. ~\cite{yu2021constructing,thiele2019spikegrad} introduced the negative spikes to cooperate with the regular positive spikes. However, the behavior of releasing negative spikes below the threshold is not consistent with the human brain. \cite{wu2021liaf} proposed the leaky integrate and analog fire neuron model to transmit the analog values among neurons, bringing performance improvements and significantly increasing energy consumption. The brain does not maintain a single spiking pattern for the same input. The coupling of different spiking patterns greatly enriches the representation ability of the spiking neurons and will adaptively cooperate to complete different cognitive functions. As the most commonly observed pattern in different brain regions, bursts might improve the selective communication between neurons~\cite{izhikevich2003bursts}, the number of spikes of the high-frequency bursts is highly robust to noise~\cite{kepecs2003information}. Although there exist some works with the burst spikes~\cite{park2019fast,li2022efficient}, their burst intensities are fixed and do not change dynamically according to the input.

\begin{figure}[h!]
      \centering
      \includegraphics[width=\linewidth]{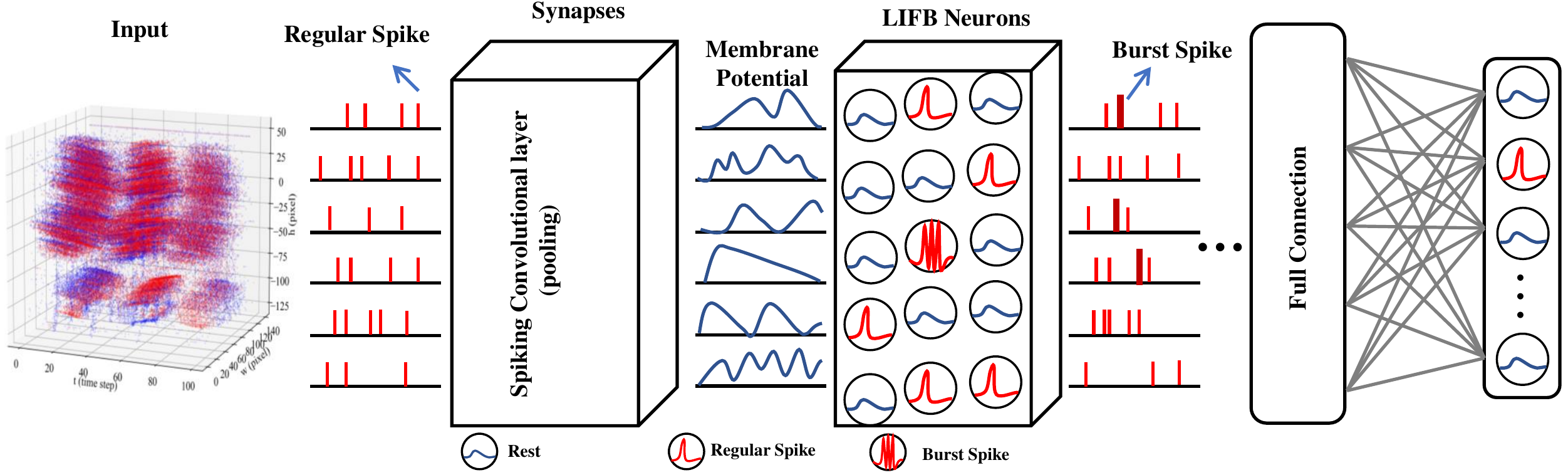}
      % Illustration of SNNs with xxx neurons
      \caption{Illustration  of the spiking neural network with our Leaky Integrate and Fire or Burst Neuron. After the network receives the input, the LIBF neurons show three spiking states: regular spike, burst spike, and resting, which significantly improve the representation ability of SNNs.}
      \label{fig:overview}
\end{figure}

In this paper, we introduce the modeling of Leaky Integrate and Fire or Burst (LIFB) neurons with three spiking patterns: resting, regular spike, and burst spike. Experiments show that our algorithm not only dramatically improves the performance of the current SNNs, but also significantly reduces the latency and energy consumption. Our contributions are summarized as follows:

\begin{itemize}
      \item We propose the Leaky Integrate and Fire or Burst neuron, as shown in Fig.~\ref{fig:overview}, which greatly improves the representation ability of the SNNs.
      \item The burst intensity of our LIFB neuron is learnable and can be dynamically adjusted according to the input.
      \item We conduct experiments on the static image datasets CIFAR10, CIFAR100, and ImageNet and the neuromorphic datasets DVS-CIFAR10 and NCALTECH101 to verify the superiority of our model. We achieve state-of-the-art performance on these datasets and achieve excellent performance using only minor simulation steps.
\end{itemize}

\section{Our Method}

\subsection{Leaky Integrate and Fire model}

The spiking neuron is the basic computational unit of SNNs. Neuroscientists have established various mathematical models such as the Hodgkin-Huxley spiking neuron (H-H)~\cite{hodgkin1952quantitative}, the Izhikevich spiking neuron~\cite{izhikevich2003simple}, Leaky Integrate and Fire spiking neuron (LIF)~\cite{dayan2005theoretical} to describe the dynamic characteristics of biological neurons. More complex mathematical models can also better describe the computational process of biological neurons. However, they also require more computational resources, while the overly complex properties are challenging to apply to the modeling of large-scale SNNs. As the most common spiking neuron model, the LIF neuron model is widely used in deep SNNs.

\begin{align}
       & \begin{aligned}
               \tau \frac{\mathrm{d}v}{\mathrm{d}t} & = -v + I,                                                     \\
                                                    & \text{if} \ v > v_{th} \ \text{, then} \ v \leftarrow v_{rst}
               \label{lif}
         \end{aligned} \\
       & s = H(v - v_{th})
      \label{spike}
\end{align}

In the Eq.~\ref{lif}, $v$ is the membrane potential, $I$ is the input current, and $v_{th}$ is the threshold. When the neuron reaches the threshold, it will deliver a spike, and the membrane potential is reset to the resting potential $v_{rst}$. $\tau$ is the membrane time constant, which controls the rate of decay of the membrane potential over time. $s$ denotes the neuronal spikes, $H(\cdot)$ denotes the heaviside step function.

The LIF model can be regarded as an integrator, capable of firing regular spikes at a constant rate and adjusting the firing rate according to the input current. To facilitate the calculation, we obtain the discrete form of Eq.~\ref{lif}:

\begin{equation}
      v(t+1) = v(t) + \frac{1}{\tau}(- v(t) + I(t))
      \label{lif_dis}
\end{equation}

Although there are many improvements for spiking neurons, they are limited to LIF neurons. The over-simplified computational characteristics of LIF neurons make it only possible to characterize regular spikes and cannot describe complex spiking patterns. There is a big gap between the LIF model and real biological neurons.

\subsection{Information Capacity for SNNs}

Spiking neuron converts continuous membrane potential into discrete spikes $s \in S = \{0, 1\}$, transmitting information through the spike train of $T$ steps. This different way of information processing from ANNs also brings differences in performance and resource costs. This paper analyzes the effect of simulation length and other properties on SNNs, and explores the relationship between SNNs and ANNs from the perspective of information capacity.

Consider a neuron with a spike train taking $t$-dimensional Boolean cubes denoted by $S = \{0, 1\}^t$. Then the effect of this neuron on a postsynaptic neuron at any step can be expressed as a (linear) threshold function on $S$ as $f: S \to \{0, 1\}$, if there exists $a \in \mathbb{R}^t$ and $b \in \mathbb{R}$:

\begin{equation}
      f(x) = H(\langle a, s \rangle + b), \ s \in S
      \label{threshold_function}
\end{equation}

The set of all threshold functions on $S$ is denoted by $T(S)$. In this way, the set of threshold functions can be used to define the capacity of the spike train:

\begin{equation}
      C(S) = \log_2{\left| T(S) \right|}
      \label{capacity}
\end{equation}

As shown in Eq.~\ref{capacity}, the capacity of a spike train is the binary logarithm of the number of all threshold functions on $S$. Considering the binary assignment associated with each partition, $C(S)$ is equivalent to the binary assignment ways the set $S$ partitioned by the hyperplane in $\mathbb{R}^n$.

The number $L(m, n)$ of connected regions created by $m$ hyperplanes through the origin in $\mathbb{R}^n$ is satisfied~\cite{winder1966partitions}:

\begin{equation}
      L(m, n) \leq 2 \sum_{k=0}^{n-1}\binom{m-1}{k}
      \label{hyperplane}
\end{equation}

According to Eq.~\ref{hyperplane}, the upper bound of $T(S)$ can be expressed as $2\sum_{k=0}^{t-1}\binom{\left|S\right|-1}{k}$.

\begin{align}
      C(S)                 & = \log_2{\left| T(S) \right|}                                                      \\
                           & \leq \log_2\left(2\sum_{k=0}^{t-1}\binom{\left|S\right|-1}{k}\right)               \\
      \label{binomial_sum} & \leq \log_2\left(2\sum_{k=0}^{t-1} \left(\frac{e\left|S\right|}{t}\right)^t\right) \\
                           & = 1 + t \log_2 \left(\frac{e\left|S\right|}{t}\right)
      \label{info_cap}
\end{align}

Eq~.\ref{binomial_sum} can be deduced from the elementary constraints on the binomial sum~\cite{vershynin2018high}.

For a spiking neuron of simulation length $t$, $\left|S\right| = 2^t$, its information capacity can be expressed as:

\begin{equation}
      C(S) \leq 1 + t^2 - t \log_2(\frac{1}{e}t)
      \label{spike_ic}
\end{equation}

ANNs use floating-point numbers to convey information. The activation value, which characterizes the output of a neuron, can also be thought of as a spike train of bits. Therefore, SNNs and ANNs with the same length have the same information capacity. In terms of information capacity, SNNs and ANNs differ only in the way they organize their spike trains. ANNs compress the spike trains into a floating-point number, while SNNs treat the spike trains as neural activity processes with temporal relationships.

By compressing the spike train into floating-point numbers, it can achieve higher training and inference efficiency on general hardware, and it can be directly trained by gradient descent, which has better performance. In contrast, establishing the temporal relationship of spikes through spiking neurons enables better temporal data processing with more complex dynamic features. However, this also makes SNNs challenging to optimize by gradient descent, resulting in performance gaps. This approach converts Multiply–Accumulate Operations (MACs) into Accumulate Operations (ACs) and has a better energy performance on neuromorphic hardware.

It is a challenging problem to organize spike sequences efficiently to achieve higher performance for a limited information capacity. Inspired by neuroscience, we found that current neuronal models can only achieve limited biological neuronal features. Biological neurons can exhibit diverse spiking patterns, which are difficult to represent by very short binary spike sequences. Therefore, combining a more biologically plausible neuron model from the perspective of information capacity, we designed an efficient neuron model by appropriately organizing the spike sequences.

\subsection{Leaky Integrate and Fire or Burst model}

Burst is a vital pattern in neurons, which contributes to gamma frequency oscillations in the brain, helps reduce neuronal noise~\cite{lisman1997bursts}, and facilitates selective communication between neurons~\cite{izhikevich2003bursts}. The Leaky Integrate and Fire or Burst (LIFB) model~\cite{smith2000fourier} is an extension of the LIF model, which retains the function of membrane potential accumulation with input current while introducing a calcium T-current parameter, bringing new properties of the burst pattern.

\begin{align}
       & \begin{aligned}
               \tau \frac{\mathrm{d}v}{\mathrm{d}t} & = -v + I + gH(v - v_h)h(v_T - v),                             \\
                                                    & \text{if} \ v > v_{th} \ \text{, then} \ v \leftarrow v_{rst}
         \end{aligned} \\
       & s = H(v - v_{th})
      \label{burst}
\end{align}

$H(\cdot)$ is the Heaviside step function. The constants $g$, $v_T$, and $v_h$ and the function $h(\cdot)$ describe the deactivation of the calcium T-current, which changes according to the membrane potential.

\begin{equation}
      \frac{\mathrm{d}h}{\mathrm{d}t} = \left\{
      \begin{aligned}
            \frac{-h}{\tau^{-}}, \ \text{if} \ v > v_h \\
            \frac{h}{\tau^{+}}, \ \text{if} \  v < v_h
      \end{aligned}
      \right.
      \label{t_current}
\end{equation}

Eq.~\ref{burst} and Eq.~\ref{t_current} describe how the T-current affects the behavior of the neuron. If the membrane potential is more significant than $v_T$ (generally $v_T > v_{rst}$), then the neuron becomes sensitive and achieves a burst mode. In contrast, Eq.~\ref{t_current} regulates the neuronal spiking pattern, and if the neuron is bursting, then the neuron becomes less sensitive to the T current and stops burst spikes.

The original LIFB model can better describe the spiking pattern of neurons, but it also brings about three times the computational costs of the LIF model~\cite{izhikevich2004model}. Meanwhile, since the current directly trained SNNs often have a small simulation step (4 or even shorter), it is not easy to show the difference between regular spikes and burst spikes, which also limits the application of the LIFB model to large-scale SNNs. In order to introduce burst mode in a small simulation step, we propose a simplified LIFB model.

\begin{align}
       & \begin{aligned}
               \tau \frac{\mathrm{d}v}{\mathrm{d}t} & = -v + I,                                                     \\
                                                    & \text{if} \ v > v_{th} \ \text{, then} \ v \leftarrow v_{rst}
         \end{aligned} \\
       & s = H(v - v_{th}) + (\kappa - 1) H(v - v_{h})
      \label{burst_s}
\end{align}

% 为了能在粗时间粒度的仿真过程中体现突发脉冲对于神经元的影响, 我们考虑了突发模式对于较短时间内对于神经元的宏观影响, 将T电流的作用叠加到神经元的输出. 为了高效地实现突发脉冲, 我们使用可训练的参数 $\kappa$ 代替Eq.~\ref{burst} 中的 $gH(v - v_h)h(v_T - v)$. 使得突发的强度能随着训练自适应地调整, 并保证简化的I\&FB高效的硬件实现. 

In order to represent the effect of burst spike on neurons in rough time simulations, we consider the macroscopic effect of burst on neurons in a shorter time, superimposing the effect of T-currents on the neuron output. To efficiently implement burst spiking, we use the parameter $\kappa$ instead of $gH(v - v_h)h(v_T - v)$ in Eq.~\ref{burst}. This allows the intensity of bursts to be adaptively adjusted while training and ensures an efficient hardware implementation of the simplified LIFB model.

\subsection{How to Represent the Burst State}

Biological neurons can transmit a large amount of information in a short period by burst spikes. The introduction of the burst spike mechanism has dramatically expanded the representation ability of LIF neurons but also brings additional computational overhead. Therefore, it is an attractive problem to define the trade-off between computational overhead and the performance of LIFB neurons.

We consider the problem of representing burst spikes from the perspective of the information capacity of neural networks. A sequence of spikes with bursts of length $t$, can be represented as $s = \{0, 1, \kappa_1, \kappa_2 \ldots, \kappa_{n - 2}\}^t$. where $n$ is the possible states at the moment. According to Eq~\ref{info_cap}, for such a sequence of spikes with burst states, the upper bound of its information capacity is:

\begin{equation}
      C(S) \leq 1 + t^2 \log_2(n) - t \log_2(\frac{1}{e}t)
      \label{info_cap_burst}
\end{equation}

\begin{figure}
      \centering  
      \subfigbottomskip=2pt 
      \subfigcapskip=-5pt 
      \subfigure[]{
            \includegraphics[width=0.48\linewidth]{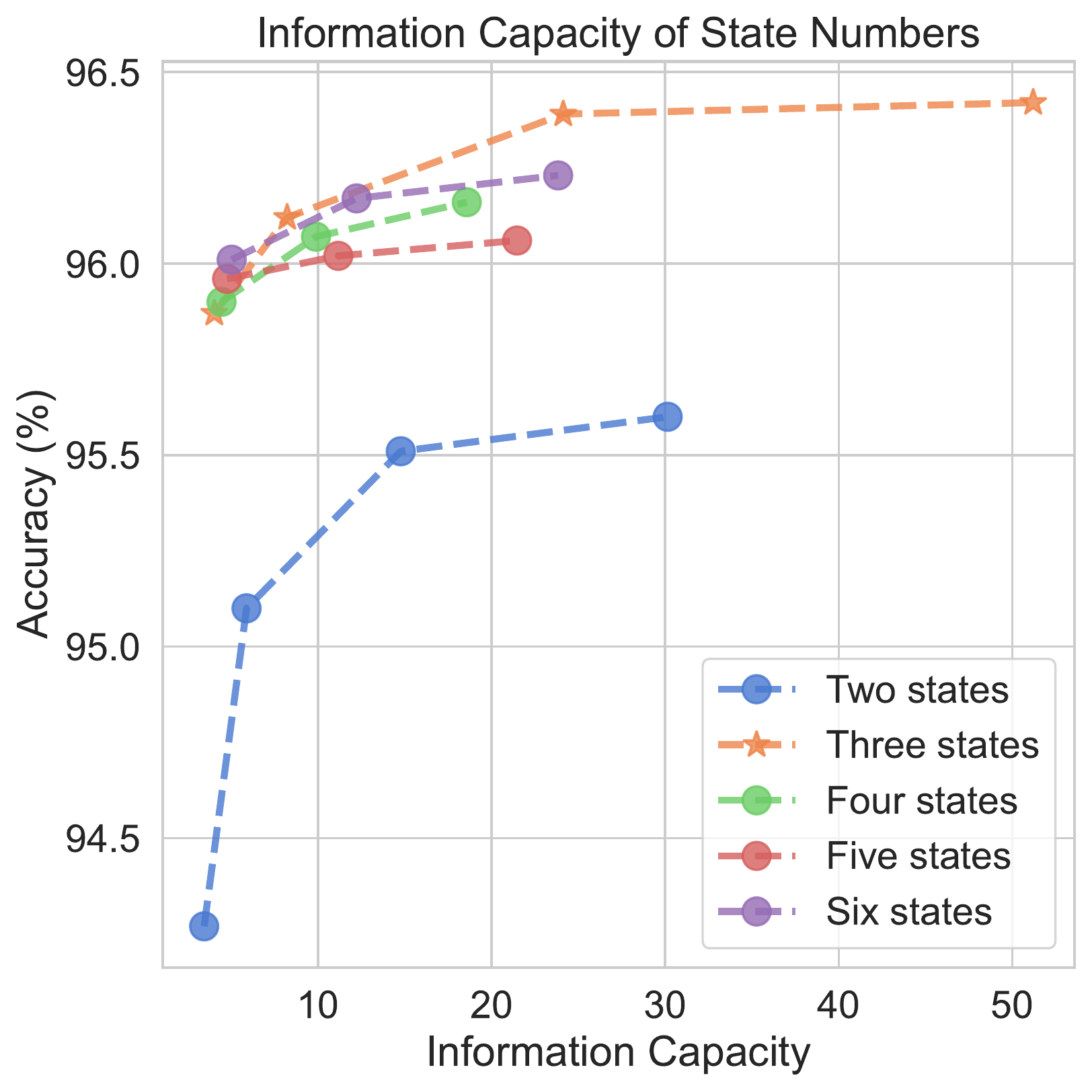}
      }\label{fig:info_cap}
      \subfigure[]{
            \includegraphics[width=0.48\linewidth]{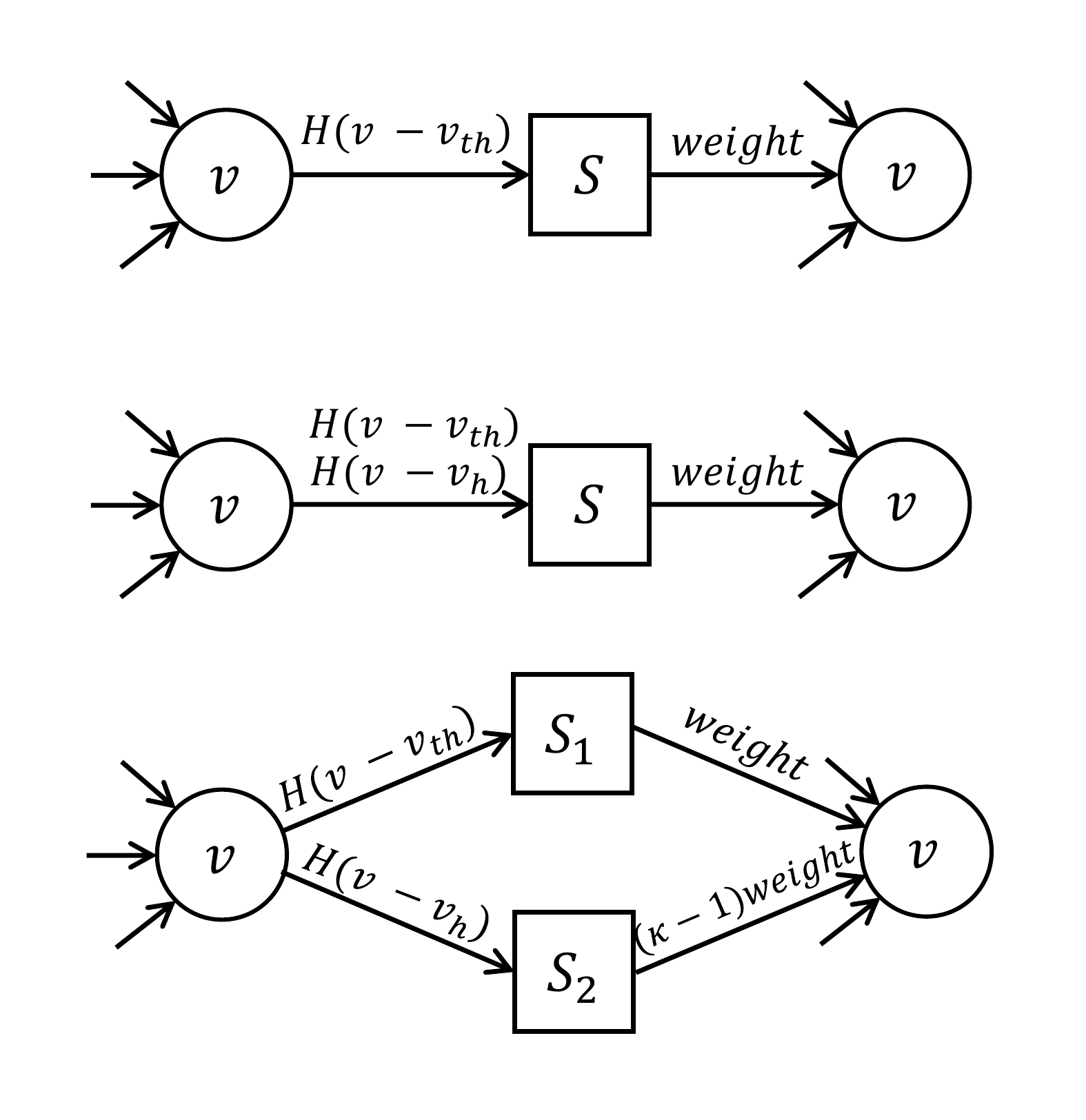}
      }\label{fig:decouple}
      \caption{(a) Relationship between information capacity and performance of number of states on CIFAR10. (b) A LIFB neuron can be decoupled into two LIF neurons. From top to bottom: LIF neuron, LIFB neuron, and LIF neuron decoupled by LIFB neuron.}
\end{figure}

As shown in Fig~\ref{fig:info_cap}, the neuron with only two states (LIF neuron) has the lowest performance. The introduction of the burst mechanism has tremendously improved the neuron's performance. It performs better if the neuron can represent multiple states at one moment. However, as the simulation time increases, too many spike states will affect the overall performance of the neuron, while $n=3$ is a trade-off between the short-time performance of the neuron and the overall performance with temporal information. Therefore, we will use $S = \{0, 1, \kappa\}$ to define the output states of the LIFB neuron.

It is worth mentioning that to introduce very few parameters and increase the support of neurons for burst spiking, we consider a learnable, channel-sharing burst intensity of $\kappa$. All neurons of the same channel use the same $\kappa$, which is negligible compared to the number of parameters of the network. To allow the burst intensity to be optimized, $\kappa$ is adjusted during the training process using gradient descent along with other parameters. We use the momentum method to ensure the stability of the burst intensity:

\begin{equation}
      \Delta \kappa := \mu \Delta \kappa + \epsilon \frac{\partial \mathcal{E}}{\partial \kappa}
      \label{optim}
\end{equation}

In Eq.~\ref{optim}, $\frac{\partial \mathcal{E}}{\partial \kappa}$ denotes the gradient propagated from the deep layer. $\mu$ is the momentum, and $\epsilon$ is the learning rate. We do not restrict the range of $\kappa$, and use $\kappa = 1$ as the initial value.

\subsubsection{Decoupling of LIFB Neurons}
\label{sec:decouple}

We present a method that can efficiently implement LIFB neurons on neuromorphic hardware that is fully compatible with existing hardware without any modification. The LIFB neuron exhibit two different spiking patterns, so it can also be decoupled into two neurons with the same input current, as shown in Fig.~\ref{fig:decouple}. This makes it easy to deploy LIFB neurons on hardware designed for LIF neurons while achieving better performance.

\section{Experiment}
In this section, we evaluate the performance of the proposed LIFB Neuron on the image datasets CIFAR10~\cite{krizhevsky2009learning}, CIFAR100~\cite{xu2015empirical} and ImageNet~\cite{russakovsky2015imagenet} and the neuromorphic datasets DVS-CIFAR10~\cite{li2017cifar10} and NCALTECH101~\cite{orchard2015converting} with BrainCog~\cite{zengBrainCogSpikingNeural2022}. The model structures used in this paper include VGG16~\cite{simonyan2014very}, ResNet20~\cite{he2016deep}, ResNet19~\cite{zheng2021going},  ResNet18-sew~\cite{fang2021deep}, and SNN6 (64C3-128C3-AP2- 256C3-AP2- 512C3-AP2-512C3-AP2-FC).

\subsection{Comparison with Other Methods}
To verify the effectiveness of our algorithm, we compare it with several current best SNNs, including conversion-based and backpropagation-based. The results for the static image data and the neuromorphic data classification task are listed in Tab.~\ref{static} and Tab.~\ref{neuromorphic}. The results for the static image dataset are at simulation steps 1, 2, 4, and 6.

\begin{table*}[t]
      \caption{Compare with existing works on static image datasets.}
      \label{static}
      % \label{sample-table}
      \resizebox{\columnwidth}{!}{
            \begin{tabular}{cccccc}
                  \hline
                  \multicolumn{1}{c}{\bf Dataset} & \multicolumn{1}{c}{\bf Model}              & \multicolumn{1}{c}{\bf Methods} & \multicolumn{1}{c}{\bf Architecture} & \multicolumn{1}{c}{\bf Simulation Length} & \multicolumn{1}{c}{\bf Accuracy} \\
                  \hline
                  \multicolumn{1}{c}{\multirow{16}{*}{CIFAR10}}
                                                  & Bu et al.~\cite{bu2021optimal}             & ANN-SNN Conversion              & ResNet-18                            & 4                                         & 90.43                            \\
                                                  & Rathi et al.~\cite{rathi2019enabling}      & Hybrid training                 & ResNet-20                            & 250                                       & 92.22                            \\
                                                  & Rathi \& Roy~\cite{rathi2020diet}          & Diet-SNN                        & ResNet-20                            & 10                                        & 92.54                            \\
                                                  & Wu et al.~\cite{wu2018spatio}              & STBP                            & CIFARNet                             & 12                                        & 89.83                            \\
                                                  & Wu et al.~\cite{wu2019direct}              & STBP NeuNorm                    & CIFARNet                             & 12                                        & 90.53                            \\
                                                  & Zhang \& Li~\cite{zhang2020temporal}       & TSSL-BP                         & CIFARNet                             & 5                                         & 91.41                            \\
                                                  & Shen et al.~\cite{shen2022backpropagation} & STBP                            & 7-layer-CNN                          & 8                                         & 92.15                            \\
                                                  & Kim et al.~\cite{kim2022neural}            & STBP                            & NAS                                  & 5                                         & 92.73                            \\
                                                  & Na et al.~\cite{na2022autosnn}             & STBP                            & NAS                                  & 16                                        & 93.15                            \\
                                                  & Zheng et al.~\cite{zheng2021going}         & STBP-tdBN                       & ResNet-19                            & 6                                         & 93.16                            \\
                                                  & Deng et al.~\cite{deng2021temporal}        & TET                             & ResNet-19                            & 6                                         & 94.50                            \\
                                                  & Guo et al.~\cite{guo2022recdis}            & Rec-Dis                         & ResNet-19                            & 6                                         & 95.55                            \\
                  \cline{2-6}
                                                  & \multirow{4}{*}{\textbf{Our Method}}       & LIFB                            & ResNet-19                            & 1                                         & \textbf{95.94}$\pm$0.09          \\
                                                  &                                            & LIFB                            & ResNet-19                            & 2                                         & \textbf{96.01}$\pm$0.07          \\
                                                  &                                            & LIFB                            & ResNet-19                            & 4                                         & \textbf{96.21}$\pm$0.10          \\
                                                  &                                            & LIFB                            & ResNet-19                            & 6                                         & \textbf{96.32}$\pm$0.06          \\
                  \hline
                  \multicolumn{1}{c}{\multirow{12}{*}{CIFAR100}}
                                                  & Bu et al.~\cite{bu2021optimal}             & ANN-SNN Conversion              & ResNet-18                            & 8                                         & 75.67                            \\
                                                  & Rathi et al.~\cite{rathi2019enabling}      & Hybrid training                 & VGG-11                               & 125                                       & 67.87                            \\
                                                  & Rathi \& Roy~\cite{rathi2020diet}          & Diet-SNN                        & ResNet-20                            & 5                                         & 64.07                            \\
                                                  & Shen et al.~\cite{shen2022backpropagation} & STBP                            & ResNet34                             & 8                                         & 69.32                            \\
                                                  & Na et al.~\cite{na2022autosnn}             & STBP                            & NAS                                  & 16                                        & 69.16                            \\
                                                  & Kim et al.~\cite{kim2022neural}            & STBP                            & NAS                                  & 5                                         & 73.04                            \\
                                                  & Deng et al.~\cite{deng2021temporal}        & TET                             & ResNet-19                            & 6                                         & 74.72                            \\
                                                  & Guo et al.~\cite{guo2022recdis}            & Rec-Dis                         & ResNet-19                            & 4                                         & 74.10                            \\
                  \cline{2-6}
                                                  & \multirow{4}{*}{\textbf{Our Method}}       & LIFB                            & ResNet-19                            & 1                                         & \textbf{77.86}$\pm$0.43          \\
                                                  &                                            & LIFB                            & ResNet-19                            & 2                                         & \textbf{78.04}$\pm$0.37          \\
                                                  &                                            & LIFB                            & ResNet-19                            & 4                                         & \textbf{78.12}$\pm$0.51          \\
                                                  &                                            & LIFB                            & ResNet-19                            & 6                                         & \textbf{78.31}$\pm$0.58          \\
                  \hline
                  \multicolumn{1}{c}{\multirow{9}{*}{ImageNet}}
                                                  & Bu et al.~\cite{bu2021optimal}             & ANN-SNN Conversion              & ResNet-34                            & 16                                        & 59.35                            \\
                                                  & Rathi et al.~\cite{rathi2019enabling}      & Hybrid training                 & ResNet-34                            & 250                                       & 61.48                            \\
                                                  & Zheng et al.~\cite{zheng2021going}         & STBP-tdBN                       & Spiking-ResNet-34                    & 6                                         & 63.72                            \\
                                                  & Deng et al.~\cite{deng2021temporal}        & TET                             & SEW-ResNet-34                        & 4                                         & 68.00                            \\
                                                  & Fang et al.~\cite{fang2021incorporating}   & SEW ResNet                      & SEW-ResNet-152                       & 4                                         & 69.26                            \\
                  \cline{2-6}
                                                  & \multirow{3}{*}{\textbf{Our Method}}       & LIFB                            & SEW-ResNet-18                        & 1                                         & \textbf{65.60}                   \\
                                                  &                                            & LIFB                            & SEW-ResNet-34                        & 1                                         & \textbf{69.34}                   \\
                                                  &                                            & LIFB                            & SEW-ResNet-34                        & 4                                         & \textbf{70.02}                   \\
                  \hline
                  % \hline
            \end{tabular}
      }
\end{table*}

\begin{table*}[t]
      \caption{Compare with existing works on neuromorphic datasets.}
      \label{neuromorphic}
      % \label{sample-table}
      \resizebox{\columnwidth}{!}{
            \begin{tabular}{cccccc}
                  \hline
                  \multicolumn{1}{c}{\bf Dataset} & \multicolumn{1}{c}{\bf Model}              & \multicolumn{1}{c}{\bf Methods} & \multicolumn{1}{c}{\bf Architecture} & \multicolumn{1}{c}{\bf Simulation Length} & \multicolumn{1}{c}{\bf Accuracy} \\
                  \hline
                  \multicolumn{1}{c}{\multirow{9}{*}{DVS-CIFAR10}}
                                                  & Zheng et al.~\cite{zheng2021going}         & STBP-tdBN                       & ResNet-19                            & 10                                        & 67.8                             \\
                                                  & Kugele et al.~\cite{kugele2020efficient}   & Streaming Rollout               & DenseNet                             & 10                                        & 66.8                             \\
                                                  & Wu et al.~\cite{wu2021liaf}                & Conv3D                          & LIAF-Net                             & 10                                        & 71.70                            \\
                                                  & Wu et al.~\cite{wu2021liaf}                & LIAF                            & LIAF-Net                             & 10                                        & 70.40                            \\
                                                  & Na et al.~\cite{na2022autosnn}             & STBP                            & NAS                                  & 16                                        & 72.50                            \\
                                                  & Shen et al.~\cite{shen2022backpropagation} & STBP                            & 5-layer-CNN                          & 16                                        & 78.95                            \\
                                                  & Guo et al.~\cite{guo2022recdis}            & Rec-Dis                         & ResNet-19                            & 10                                        & 72.42                            \\
                                                  & Deng et al.~\cite{deng2021temporal}        & TET                             & VGGSNN                               & 10                                        & 83.17                            \\
                                                  & \textbf{Our Method}                        & LIFB                            & SNN7                                 & 10                                        & \textbf{83.83}$\pm$0.70          \\
                  \hline
                  \multicolumn{1}{c}{\multirow{3}{*}{N-Caltech101}}
                                                  & Kugele et al.~\cite{kugele2020efficient}   & STBP                            & VGG11                                & 20                                        & 55.0                             \\
                                                  & Ramesh et al.~\cite{ramesh2019dart}        & N/A                             & N/A                                  & N/A                                       & 66.8                             \\
                                                  & \textbf{Our Method}                        & LIFB                            & SNN7                                 & 10                                        & \textbf{81.74}$\pm$0.81          \\
                  \hline
            \end{tabular}
      }
\end{table*}

For CIFAR10 and CIFAR100, LIFB achieves higher accuracy than previous work at a simulation length of $1$. In particular, using the same network structure and simulation step length, our LIFB also has a significant advantage, improving $0.59\%$ and $3.31\%$ on CIFAR10 and CIFAR100 compared with  Rec-Dis~\cite{guo2022recdis}.

For the more challenging ImageNet dataset, we achieve $65.60\%$ accuracy using only the lightweight ResNet18 structure. Moreover, we achieve better performance than SEW-ResNet152.~\cite{fang2021incorporating} when using the SEW-ResNet34 structure only at a simulation length of $1$. Our LIFB achieves $2.02\%$ improvements using the same structure and simulation length as previous work.

For the neuromorphic dataset DVS-CIFAR10, our LIFB achieves state-of-the-art performance by using only the SNN7 structure with less than half the parameters of VGGSNN~\cite{deng2021temporal}. For the N-Caltech101 dataset, we achieved $83.44\%$ top-1 accuracy, achieving a performance far beyond previous work.

\begin{figure}[h!]
      \centering
      \includegraphics[width=.9\linewidth]{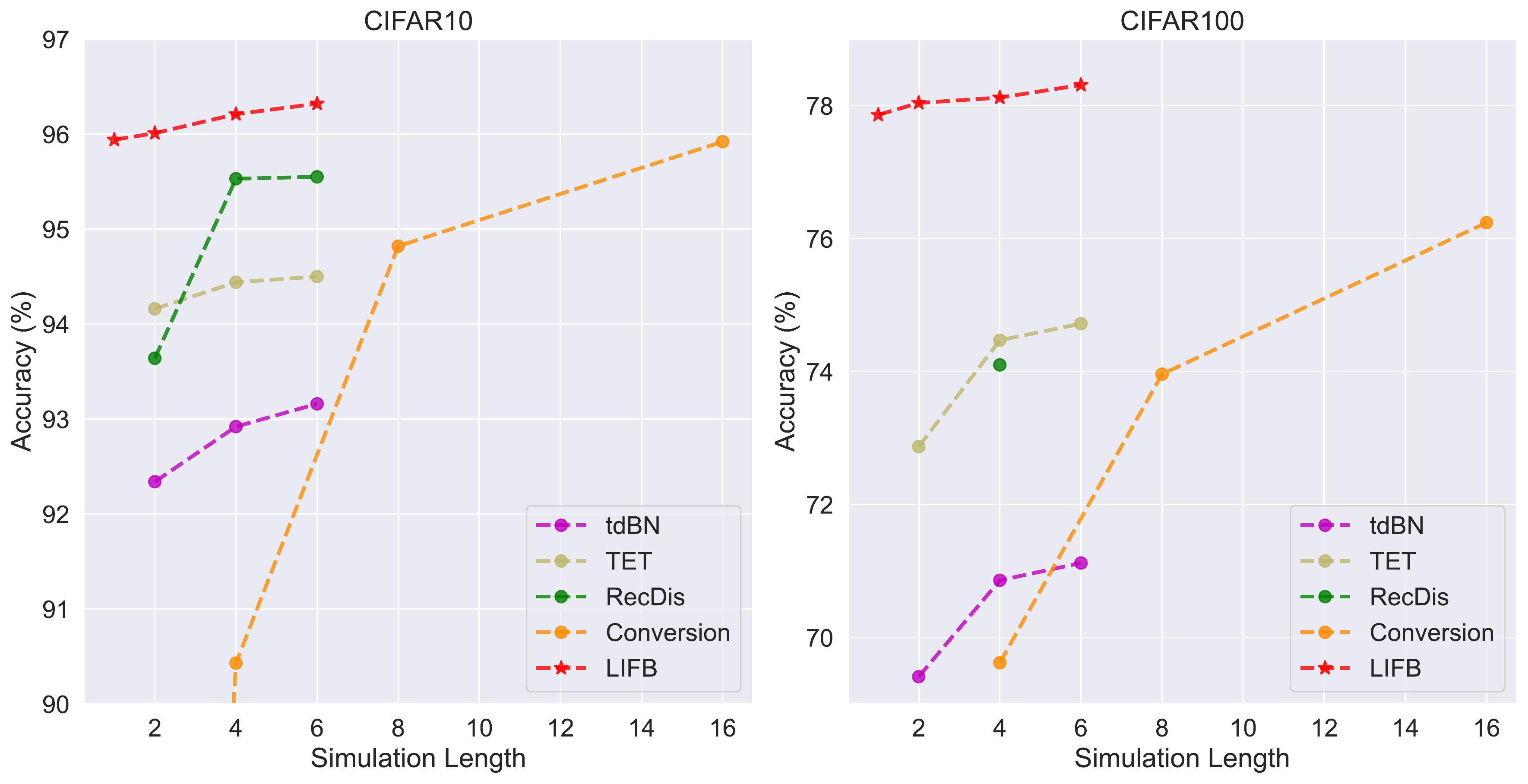}
      % CIFAR10/100数据集上仿真长度和准确率的关系.
      \caption{Relationship between simulation length and accuracy on the CIFAR10/100 dataset.}
      \label{step_acc}
\end{figure}

To further illustrate the advantages of our LIFB, we show the comparison with previous methods at different simulation lengths. As shown in Fig.~\ref{step_acc}, we compared LIFB with directly trained SNNs and converted SNNs. Our LIFB shows a significant advantage at shorter simulation lengths due to its more vital representation ability.

\subsection{Ablation Studies}

Compared with other advanced methods, LIFB allows models to exhibit better performance and achieve better top-1 accuracy on classification tasks. We conducted ablation studies to further verify the contribution of LIFB for different network structures and simulation lengths.

\begin{table*}[h]
      \centering
      \caption{Ablation studies on different network structures and simulation lengths.}
      % \resizebox{\linewidth}{!}{
      \begin{tabular}{cccccccccc}
            \toprule[1pt]
            \multirow{2}{*}{Architecture} & \multirow{2}{*}{Neuron} & \multicolumn{4}{c}{CIFAR10} & \multicolumn{4}{c}{CIFAR100}                                                 \\
            \cmidrule(r){3-6}  \cmidrule(r){7-10}
                                          &                         & 1                           & 2                            & 4     & 6     & 1     & 2     & 4     & 6     \\
            \hline
            \multirow{2}{*}{VGG16}        & LIF                     & 92.02                       & 93.41                        & 94.13 & 94.08 & 66.40 & 69.18 & 71.15 & 71.99 \\
                                          & LIFB                    & 94.53                       & 95.02                        & 95.28 & 95.36 & 73.00 & 73.90 & 74.34 & 75.02 \\
            \hline
            \multirow{2}{*}{ResNet19}     & LIF                     & 93.76                       & 94.44                        & 95.07 & 95.51 & 73.70 & 74.34 & 75.01 & 75.62 \\
                                          & LIFB                    & 95.94                       & 96.01                        & 96.21 & 96.32 & 77.86 & 78.04 & 78.12 & 78.31 \\
            \hline
            \multirow{2}{*}{ResNet20}     & LIF                     & 83.65                       & 86.47                        & 88.09 & 89.16 & 49.93 & 53.90 & 57.40 & 58.17 \\
                                          & LIFB                    & 89.72                       & 90.88                        & 91.30 & 91.65 & 60.63 & 62.95 & 63.28 & 64.33 \\
            \hline
            \multirow{2}{*}{SEW-ResNet18} & LIF                     & 94.27                       & 95.10                        & 95.51 & 95.60 & 72.59 & 74.16 & 75.68 & 76.52 \\
                                          & LIFB                    & 95.87                       & 96.12                        & 96.39 & 96.42 & 75.88 & 77.38 & 78.41 & 78.67 \\
            \bottomrule[1pt]
      \end{tabular}
      % }
      \label{tab:ablation}
\end{table*}

\begin{figure}[h!]
      \centering
      \includegraphics[width=1.\linewidth]{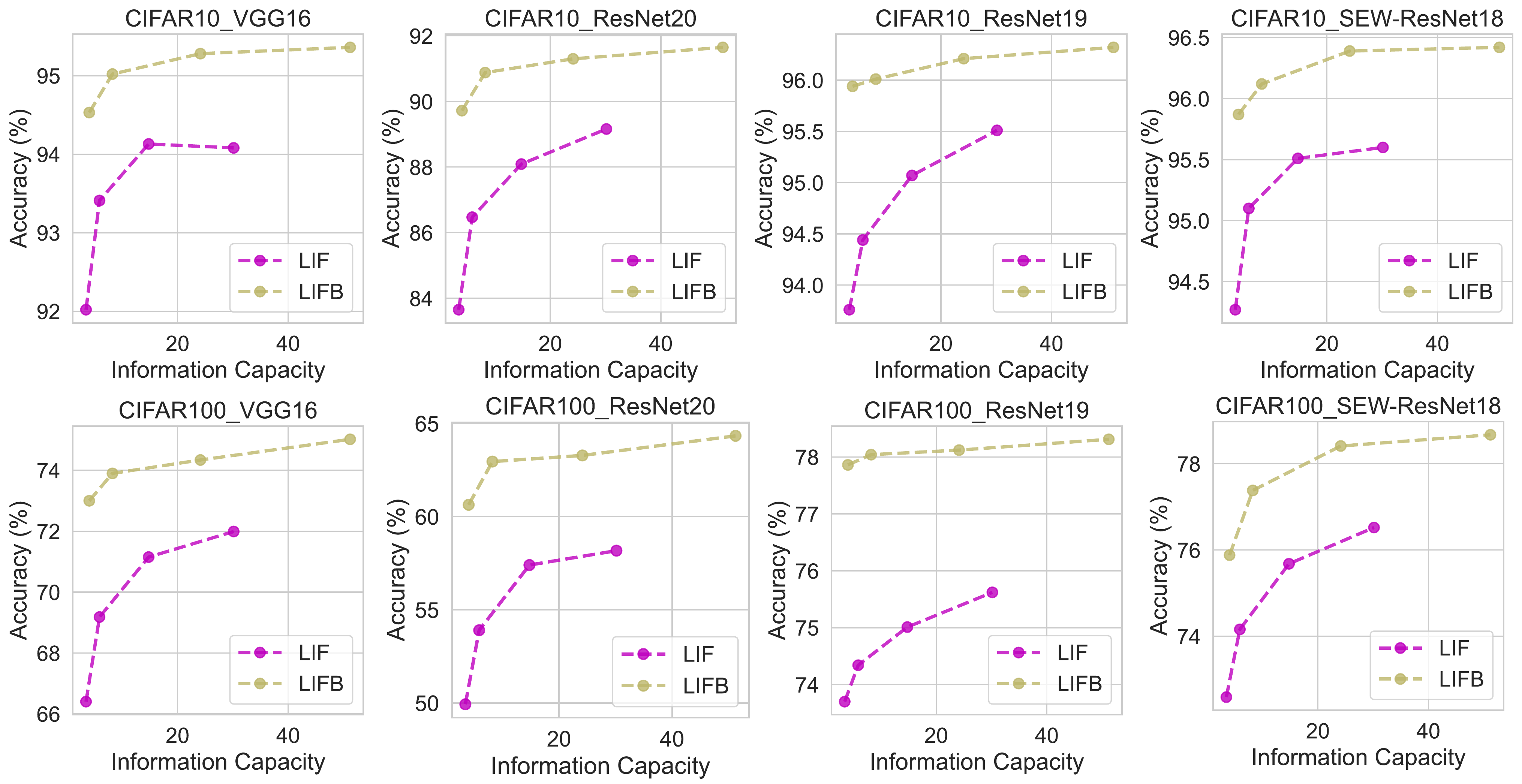}
      \caption{Effect of neuron type on information capacity and performance.}
      \label{fig:ablation}
\end{figure}

As shown in Tab.~\ref{tab:ablation}, LIFB maintained its advantage over LIF neurons for all models and all simulation lengths. LIFB also has higher accuracy than LIF neurons with longer simulation time only at the simulation length $t=1$. Fig.~\ref{fig:ablation} shows a comparison of the impact of neuron type in terms of the information capacity of neurons. It can be seen that our LIFB still maintains a higher accuracy than LIF with the same information capacity.

\begin{figure}[h!]
      \centering
      \includegraphics[width=1.\linewidth]{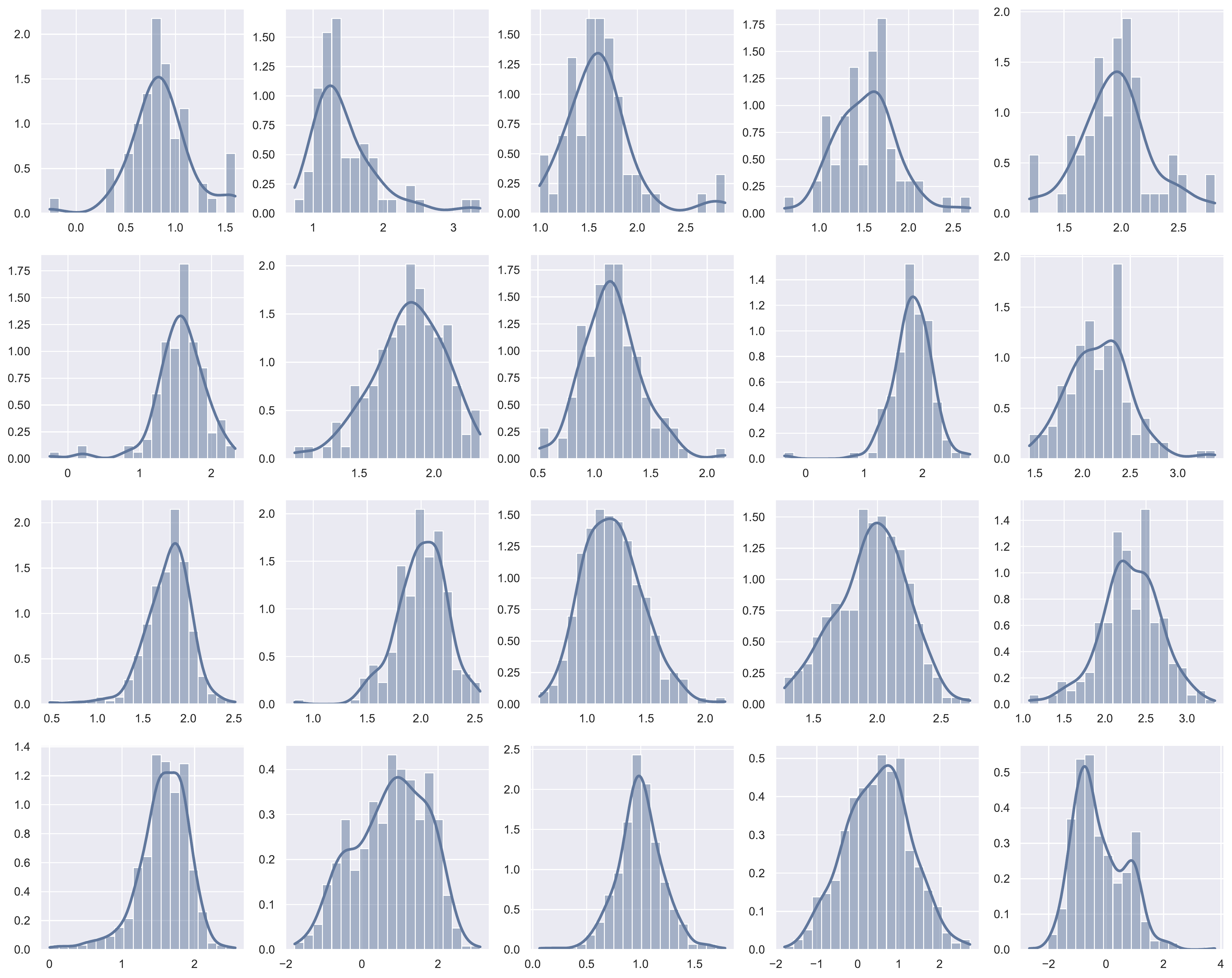}

      \caption{Distribution of burst intensity of well-trained SEW-ResNet18 on CIFAR10.}
      \label{fig:burst_intensity}
\end{figure}

In the original LIFB, the switching of neuronal spiking patterns is achieved by adjusting the conductance of the calcium T-current. However, this slow adjustment is challenging to be effective at short simulation lengths and is also difficult to be applied to large-scale SNNs due to the high computational costs. Therefore, we propose a simplified LIFB neuron. We directly apply the effect of T-current conductance to the neuron output and optimize the burst intensity by the learnable parameter $\kappa$. This channel-sharing burst intensity ensures flexible spiking characterization with very few additional parameters. Fig~\ref{fig:burst_intensity} shows the distribution of burst intensity of different layers of the SEW-ResNet18 trained on CIFAR10.

We conducted an ablation study on learnable burst intensity, as shown in Tab.~\ref{tab:intensity}. We compared the top-1 accuracy of neurons with fixed burst intensity and neurons with learnable synaptic strengths on the CIFAR10 dataset.

\begin{table}[h]
      \centering
      \caption{Comparison of fixed/learnable burst intensity of VGG16 on CIFAR10.}
      % \resizebox{\linewidth}{!}{
      \begin{tabular}{ccccccc}
            \hline
            $\kappa$ & LIF   & 0.5   & 1     & 1.5   & 2     & learnable      \\
            \hline
                     & 92.02 & 92.13 & 94.17 & 94.11 & 93.52 & \textbf{94.53} \\
            \hline
      \end{tabular}
      % }
      \label{tab:intensity}
\end{table}

The learnable channel-sharing burst intensity enables neurons to learn the appropriate burst from the data, and neurons can achieve bursts of arbitrary strength compared to regular spikes, which is more biologically plausible and enhances the performance of SNNs.

\subsection{Comparison with Other Triple-value Neurons}

By considering the different spiking patterns of neurons, we design LIFB neurons that can exhibit triple neuronal states: rest, regular spike, and burst spike. Our LIFB neuron expands the representation ability of neurons, is more biologically plausible, and exhibits better energy efficiency and performance than other binary-value neurons.

In addition to our LIFB neuron, there are many works that enable neurons to fire positive and negative spikes (PosNeg) to achieve triple-value representation~\cite{yu2021constructing,thiele2019spikegrad}. Here we consider the same approach.

We compare our LIFB neuron with the PosNeg at different simulation lengths on CIFAR10 dataset as shown in Tab.~\ref{tab:exc_inh}. Our LIFB neuron shows better performance at different simulation lengths.

\begin{table}[h]
      \centering
      \caption{Comparison of LIFB neuron with PosNeg neuron on CIFAR10.}
      % \resizebox{\linewidth}{!}{
      \begin{tabular}{ccccc}
            \hline
            Neuron        & 1     & 2     & 4     & 6     \\
            \hline
            PosNeg        & 93.87 & 94.16 & 94.33 & 94.43 \\
            \textbf{LIFB} & 94.53 & 94.91 & 95.17 & 95.15 \\
            \hline
      \end{tabular}
      % }
      \label{tab:exc_inh}
\end{table}

\subsection{Loss Landscape around Local Minima}

\begin{figure}[h!]
      \centering
      \includegraphics[width=0.9\linewidth]{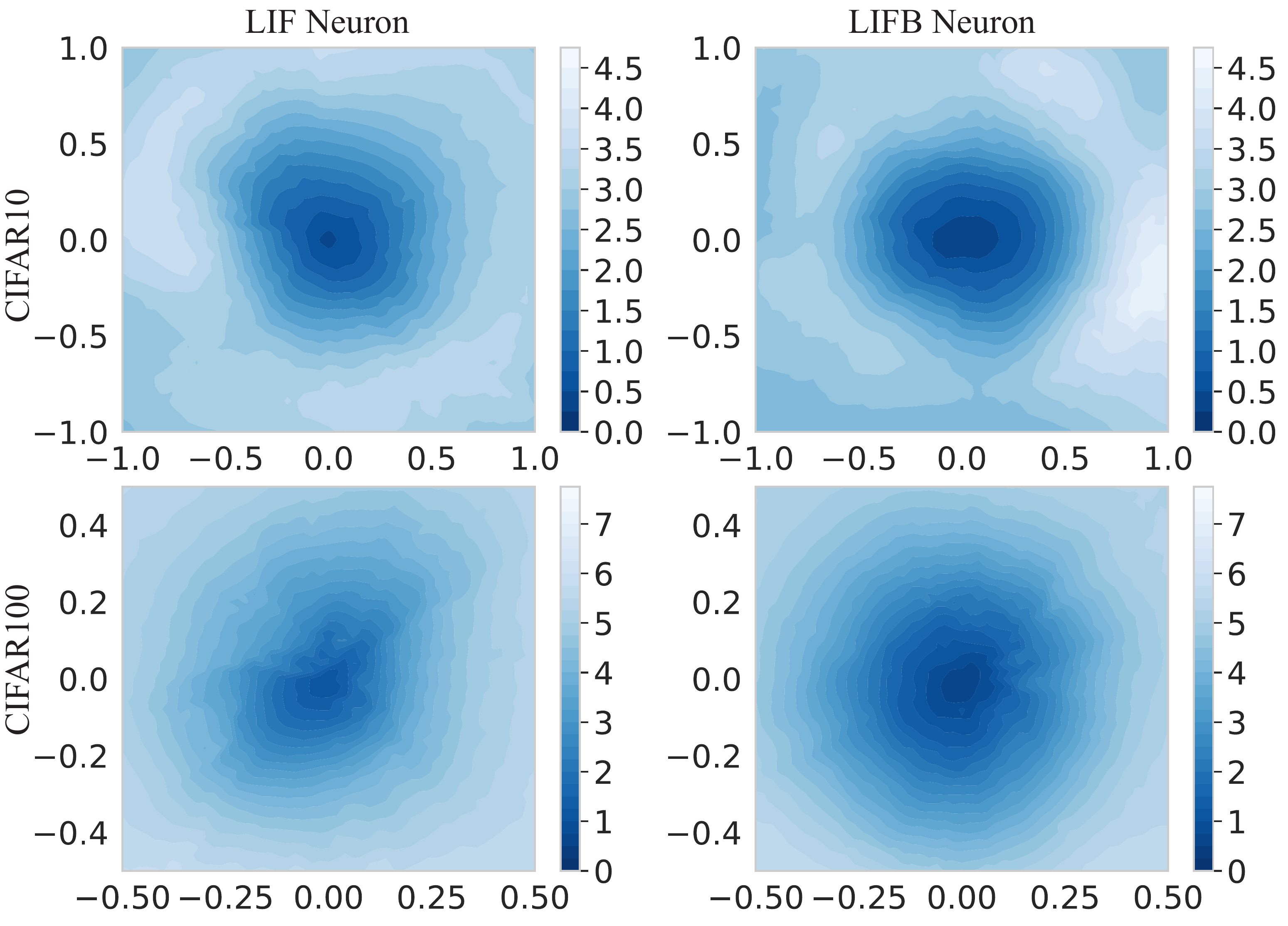}
      % 不同神经元VGG16上损失景观的对比
      \caption{Comparison of loss landscapes of different neurons on VGG16.}
      \label{fig:landscape}
\end{figure}

We further show the 2D landscapes of SNNs with different types of neurons around their local minima \cite{li2018visualizing} to verify the effect of LIFB neurons on the generalization ability. As shown in Fig.~\ref{fig:landscape}, we show the local 2D landscape of the VGG16 model on CIFAR10/100, using different neurons. It can be seen that LIFB Neuron finds flatter local minima and more minor losses. This further demonstrates the ability of the LIFB neuron to enhance the representation and generalization of the model.

\subsection{Comparison of LIFB with decoupled LIF}

LIFB neuron achieves much better performance than LIF Neuron by better modeling biological neurons, but this also entails additional computational overhead. Although the above experiments have demonstrated that LIFB neurons perform better than LIF neurons with longer simulation times, this cannot be achieved by increasing computational resources. To further illustrate that the performance gain of LIFB neurons comes from more reasonable modeling rather than more computational resources, we performed a fairer comparison.

As discussed in Sec.~\ref{sec:decouple}, a LIFB neuron can be decoupled into two LIF neurons with the same input current and different threshold voltages. We, therefore, compared the LIFB Neuron with its equivalent decoupled LIF neuron trained from scratch, as shown in Tab.~\ref{tab:decouple}. The results directly indicate that most of the performance gains from LIFB neurons come from well-formulated spiking pattern design rather than from the higher computational costs.

\begin{table}[h]
      \centering

      \caption{Comparison of LIFB neurons and equivalent decoupled LIF neurons trained from scratch.}

      \begin{tabular}{ccccc}
            \hline
            Neuron        & 1     & 2     & 4     & 6     \\
            \hline
            LIF           & 92.02 & 93.41 & 94.13 & 94.08 \\
            Scratch       & 93.78 & 94.36 & 94.83 & 95.02 \\
            \textbf{LIFB} & 94.53 & 95.02 & 95.28 & 95.36 \\
            \hline
      \end{tabular}
      \label{tab:decouple}
\end{table}

\subsection{Visualization of Neural Activity}

\begin{figure}[!htbp]
      \centering
      \includegraphics[width=1.\linewidth]{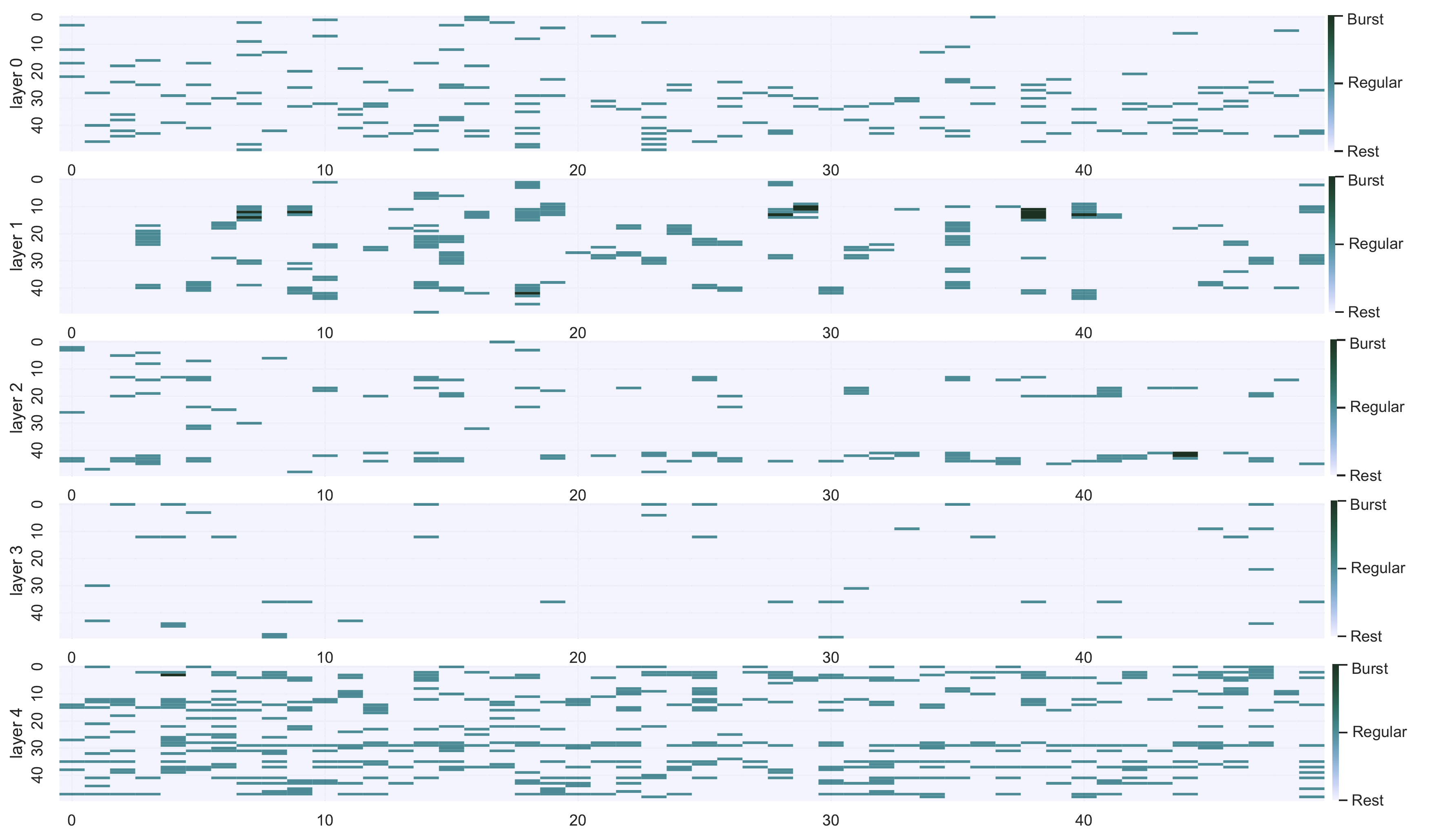}
      \caption{Visualization of neural activity of LIFB neurons on CIFAR10 dataset of VGG7.}
      \label{fig:vis}
\end{figure}

The neural activity of VGG7 on CIFAR10 dataset with LIFB neurons at different layers is shown in Fig.~\ref{fig:vis}. We randomly selected 50 neurons in each layer, with cyan color indicating that the neuron is at a regular spike state and dark cyan indicating that the neuron is at a burst spike state. Although neurons are rarely in burst mode, this biologically plausible neuron model is essential for the performance of SNNs.

\section{Conclusion}
Inspired by the multi-spike delivery form of the brain, we design an efficient Leaky Integrate and Fire or Burst neuron model with triple-valued output from the perspective of network information capacity, while the burst density in LIFB can be adaptively adjusted. This multi-spike issuing form of synergistic neurons greatly enriches the characterization capability of the SNNs. Experimental results on static datasets CIFAR10, CIFAR100, and ImageNet show that we only need one simulation step to achieve a very high accuracy, which significantly reduces the latency of the SNNs. Also, we achieve state-of-the-art performance on the neuromorphic datasets DVS-CIFAR10 and NCALTECH101.

\bibliography{refs}
\bibliographystyle{icml2021}

\end{document}